\documentclass[10pt,twocolumn,letterpaper]{article}

\usepackage{cvpr}
\usepackage{times}
\usepackage{epsfig}
\usepackage{graphicx}
\usepackage{amsmath}
\usepackage{amssymb}

\usepackage[pagebackref=true,breaklinks=true,letterpaper=true,colorlinks,bookmarks=false]{hyperref}

\cvprfinalcopy 

\def\cvprPaperID{2} 

\ifcvprfinal\fi
\begin{document}

\title{Object-Scene Convolutional Neural Networks for Event Recognition in Images }


\author{Limin Wang$^{1,2}$ \quad \quad Zhe Wang$^{2}$ \quad \quad Wenbin Du$^{2}$ \quad \quad Yu Qiao$^{2}$ \\
\small $^{1}$Department of Information Engineering, The Chinese University of Hong Kong\\
\small $^{2}$Shenzhen key lab of Comp. Vis. \& Pat. Rec.,  Shenzhen Institutes of Advanced Technology, CAS, China \\
{\tt\small 07wanglimin@gmail.com, buptwangzhe2012@gmail.com, wb.du@siat.ac.cn, yu.qiao@siat.ac.cn}
}

\maketitle

\begin{abstract}
Event recognition from still images is of great importance for image understanding. However, compared with event recognition in videos, there are much fewer research works on event recognition in images. This paper addresses the issue of event recognition from images and proposes an effective method with deep neural networks. Specifically, we design a new architecture, called \emph{Object-Scene Convolutional Neural Network} (OS-CNN). This architecture is decomposed into object net and scene net, which extract useful information for event understanding from the perspective of objects and scene context, respectively. Meanwhile, we investigate different network architectures for OS-CNN design, and adapt the deep (AlexNet) and very-deep (GoogLeNet) networks to the task of event recognition. Furthermore, we find that the deep and very-deep networks are complementary to each other. Finally, based on the proposed OS-CNN and comparative study of different network architectures, we come up with a solution of five-stream CNN for the track of cultural event recognition at the ChaLearn Looking at People (LAP) challenge 2015. Our method obtains the performance of $85.5\%$ and ranks the $1^{st}$ place in this challenge.
\end{abstract}

\section{Introduction}
Event recognition from still images is one of the challenging problems in computer vision research. While many efforts have been devoted to the problem of video-based event and action recognition \cite{SimonyanZ14b,SunN13,WangS13a,WangQT13b,WangQT13a,WangQT15a}, there are much fewer research works on image-based event recognition \cite{LiF07,XiongZLT15}. Compared with images, videos are able to provide more useful information for event undertanding such as motion, in addition to static appearance. Therefore, event recognition from still images poses more challenges than videos. Meanwhile, the concept of event itself is extremely complex, as the characterization of an event is related to many factors, including objects, human poses, human garments, scene categories, and other context. Therefore, the event recognition is highly related with other high-level computer vision problems, such as object recognition \cite{KrizhevskySH12} and scene recognition \cite{ZhouLXTO14}. In this paper, we propose an effective method for the track of cultural event recognition at the ChaLearn Looking at People (LAP) challenge 2015 \cite{LAP15}, which obtains the performance of $85.5\%$ and ranks the $1^{st}$ place in this challenge.

Specifically, we propose a new architecture for event recognition, called \emph{Object-Scene Convolutional Neural Network} (OS-CNN), which extracts the important visual cues of both object and scene for event understanding. The OS-CNN is decomposed into two separate nets, namely object net and scene net. The object net aggregates important information for recognizing event from the perspective of object, while the scene net performs event recognition with the help of scene context. The cues of containing object and scene context provide complementary information for event understanding from still images. The recognition results from object net and scene net are combined by late fusion. Decoupling the object and scene nets also allows us to exploit the availability of large amounts of annotated image data by pre-training object net on the ImageNet challenge dataset \cite{DengDSLL009} and scene net on the Places dataset \cite{ZhouLXTO14}.

Meanwhile, there are many famous and successful network architectures for CNNs, such as AlexNet \cite{KrizhevskySH12}, ClarifaiNet \cite{ZeilerF14}, GoogLeNet \cite{SzegedyLJSRAEVR14}, and VGGNet \cite{SimonyanZ14a}. These architectures have proved to be effective for object and scene recognition, and obtained the state-of-the-art performance on the datasets of ImageNet and Places \cite{ILSVRCarxiv14,ZhouLXTO14}. However, their performance on event recognition and the complementarity among them has not been fully explored before. In our proposed OS-CNN, we exploit these successful deep architectures for event recognition, and further boost the recognition performance by using ensemble of them. Finally, based on our OS-CNN and comparative study of different network architectures, we come up with a solution of five-stream CNN for the ChaLearn LAP challenge 2015.

\begin{figure*}
  \includegraphics[width=\textwidth]{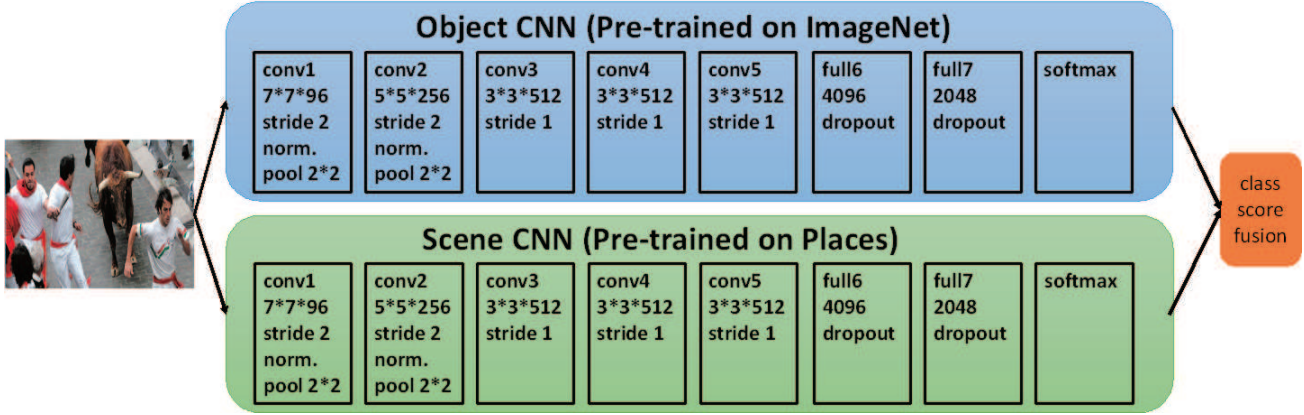}
  \caption{The architecture of Object-Scene Convolutional Neural Network (OS-CNN) for event recognition. We pre-trained the object CNN on the ImageNet dataset and the scene CNN on the Places dataset. It should be noted that we choose the ClarifaiNet architecture for CNN in this illustration. But, in practice, we may choose other architectures for both object and scene CNN and even fuse  multiple different architectures.}
  \label{fig:pipeline}
\end{figure*}

The rest of this paper is organized as follows. In Section \ref{sec:oscnn}, we describe the technical details about our OS-CNN. We then provide the implementation details and experimental results in Section \ref{sec:result}. Finally, we conclude our method and present the future works in Section \ref{sec:conclusion}.

\section{Object-Scene CNNs}
\label{sec:oscnn}

Event understanding is highly related with other two high-level computer vision problems: object and scene recognition. Therefore, we utilize two separate components for event recognition. The object stream, pre-trained in large object dataset (ImageNet), carries information about object depicted in the image. The scene stream, pre-trained in large scene dataset (Places), captures the pattern about scene context of this image. We design our event recognition architecture accordingly and propose a new network architecture, called Object-Scene CNN (OS-CNN) as shown in Figure \ref{fig:pipeline}. Each CNN is pre-trained on its own dataset and fine tuned for event recognition on the target dataset. We use late fusion to combine the scores of two separate CNNs.

\subsection{Object nets}

We hope that the object net is able to capture useful information of object to help event recognition. As the object net is essentially dealing with object cues, we build it with the help of the recent advances on large-scale image recognition methods \cite{KrizhevskySH12}, and pre-train the network on a large image classification dataset, such as the ImageNet dataset \cite{DengDSLL009}. Specifically, we first choose the ClarifaiNet network architecture \cite{ZeilerF14} and use the pre-trained model in \cite{ChatfieldSVZ14} \footnote{\url{http://www.robots.ox.ac.uk/~vgg/software/deep_eval/}}. Then, we fine tune the model parameters for the task of event recognition on the training dataset provided by the challenge organizers. The details about the network architecture can be referred to its original paper \cite{ZeilerF14} and the details about the fine tuning of network parameters can be found in Section \ref{sec:result}. Next, we describe the scene net, which exploits scene information for event recognition.

\subsection{Scene nets}

The scene net is expected to extract the scene information of image to help conduct event recognition. Hence, the scene net is designed for handling scene context, and we may resort to recent advances on the problem of scene recognition. Places dataset \cite{ZhouLXTO14} is a recent large dataset and it contains $205$ scene categories with $2.5$ millions of images. Specifically, we first use the pre-trained model in \cite{ZhouLXTO14} \footnote{\url{http://places.csail.mit.edu/}}, which choose the famous AlexNet architecture \cite{KrizhevskySH12}. Similar to object net, we then fine tune the model parameters on the training dataset from the cultural event recognition challenge. The details about the network architecture can be found in its original paper \cite{KrizhevskySH12} and the details about the fine tuning of network parameters can be found in Section \ref{sec:result}.

Based on the above analysis, the recognition of event is highly related to the concepts of object and scene. Therefore, we expect that the prediction results of both object and scene nets are complementary to each other, and combine them using late fusion as follows:
\begin{equation}
  s(\mathbf{I}) = \alpha_o s_o(\mathbf{I}) + \alpha_s s_s(\mathbf{I}),
  \label{equ:fusion1}
\end{equation}
where $\mathbf{I}$ is the input image, $s_o(\mathbf{I})$ and $s_s(\mathbf{I})$ are the prediction scores of object and scene net, $\alpha_o$ and $\alpha_s$ are the fusion weights of object and scene net. In current implementation, these two fusion weights are equal to each other.

\subsection{Ensemble of multiple CNNs}
In the past several years, several successful deep CNN architectures have been designed for the task of object recognition at the ImageNet Large Scale Visual Recognition Challenge \cite{ILSVRCarxiv14}. These architectures can be roughly classified into two categories: (i) deep CNN including AlexNet \cite{KrizhevskySH12} and ClarifaiNet \cite{ZeilerF14}, (ii) very-deep CNN including GoogLeNet \cite{SzegedyLJSRAEVR14} and VGGNet \cite{SimonyanZ14a}. The deep CNN architectures usually contain $5$ convolutional layers and $3$ fully-connected layers as shown in Figure \ref{fig:pipeline}. The very-deep CNN architectures resort to extremely deep structures with smaller initial filter size or designing a new inception module in a network-in-network manner. The previous studies show that deeper networks will obtain better performance on the task of object recognition. However, their performance on event recognition still remains unknown.

In this subsection, we exploit these very-deep networks in our proposed Object-Scene CNN architecture and aim to verify the superior performance of deeper structure. Specifically, we choose the GoogLeNet architecture for both object and scene nets. GoogLeNet is a 22-layer very-deep network and is based on a newly-designed module, codenamed \emph{Inception}. To optimize performance, the architectural decisions are based on the Hebbian principle and the intuition of multi-scale processing. The details about GoogLeNet architecture can be found in \cite{SzegedyLJSRAEVR14}. We use the GoogLeNet model released on the Caffe webpage \footnote{\url{https://github.com/BVLC/caffe/wiki/Model-Zoo}} to initialize the object net. For scene CNN, we utilize the pre-trained model released in the technical report \cite{WuZYX14} \footnote{\url{http://vision.princeton.edu/pvt/GoogLeNet/}}.

We also study the complementarity of convolutional neural networks with different architectures. We combine the prediction results of deep OS-CNNs with the ones of very-deep OS-CNNs as follows:
\begin{equation}
  s_x(\mathbf{I}) = \beta^{\mathrm{d}} s_x^d(\mathbf{I}) + \beta^{\mathrm{v-d}} s_x^{v-d}(\mathbf{I}),
  \label{equ:fusion1}
\end{equation}
where $x \in \{o,s\}$ denotes object net or scene net, $s_x^d(\mathbf{I})$ and $s_x^{v-d}(\mathbf{I})$ are the scores of deep and very-deep CNNs, $\beta^{\mathrm{d}}$ and $\beta^{\mathrm{v-d}}$ are their fusion weights ($\beta^{\mathrm{d}} = 0.3$ and $\beta^{\mathrm{v-d}} = 0.6$). Although the very-deep OS-CNN outperforms the deep OS-CNN, the combination of them is still able to further boost the recognition performance.

\section{Experiments}
\label{sec:result}
\begin{figure*}[t]
  \includegraphics[width=\textwidth]{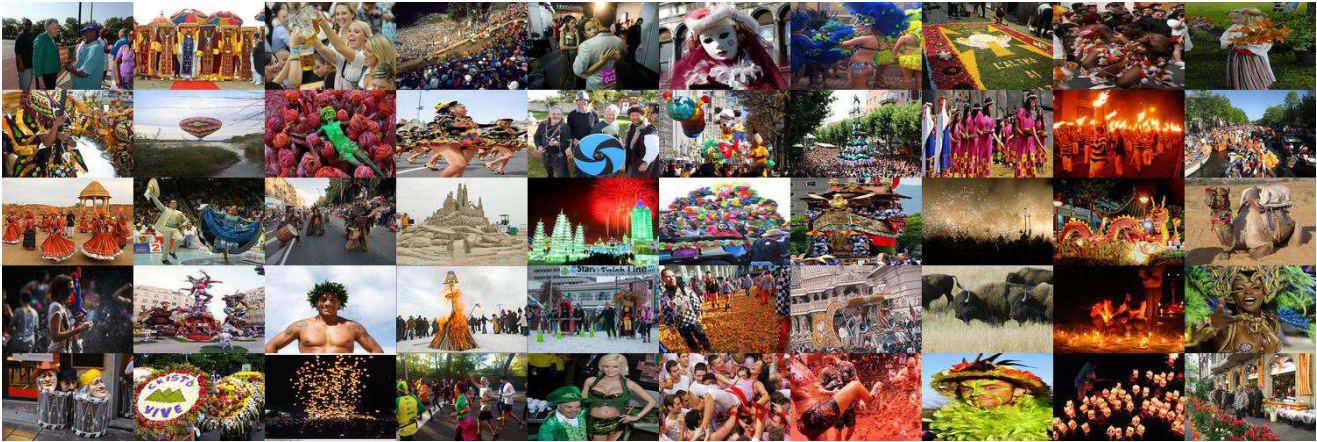}
  \caption{Samples of cultural event recognition dataset at the ChaLearn LAP challenge 2015. The cultural event recognition dataset has $50$ important cultural events in the world. It includes: Annual Buffalo Roundup (USA), Battle of the Oranges (Italy), Chinese New Year (China), Notting Hill Carnival (UK), Obon (Japan) and so on. All the images are collected from the Internet by using Google and Bing search engines. These images exhibit large intra-class variations and are very challenging for event recognition.}
  \vspace{-0.4cm}
  \label{fig:sample}

\end{figure*}

In this section, we first describe the dataset of cultural event recognition at the ChaLearn LAP challenge 2015. Then we give a detailed description of the implementation details about training OS-CNNs on the event recognition dataset provided by the challenge organizers. Finally, we present and analyze the experimental results of proposed OS-CNNs on the dataset of ChaLearn LAP challenge 2015.

\subsection{Datasets and evaluation protocal}
Cultural event recognition is a new task at the ChaLearn LAP challenge 2015. This task provides an event recognition dataset composed of images collected from two image search engines (Google Images and Bing Images). There are $50$ important cultural events from the world in this dataset, and some sample images are shown in Figure \ref{fig:sample}. From these images, we see that garments, human poses, objects, and scene context constitute the possible cues to be exploited for recognizing the events. The dataset is divided into three parts: development data ($5,875$ images), validation data ($2,332$ images), and evaluation data ($3,569$ images). During develop phase, we train our model on the development data and verify its performance on the validation data. For final evaluation, we merge the development and validation data into a single training data ($8,207$ images), and re-train our model. Our final submission results to the challenge are obtained by the re-trained model. The principal quantitative measure used is based on  precision/recall curve. They use the area under this curve as the computation of the average precision (AP), which is calculated by numerical integration.

\subsection{Implementation details}
The training procedure of OS-CNNs is implemented using the famous Caffe toolbox \cite{JiaSDKLGGD14}. Although there are $8,207$ training images in the cultural event recognition dataset, its size is relatively small compared with the ImageNet dataset \cite{DengDSLL009}. Therefore, we choose to pre-train our model on two large datasets: ImageNet dataset for object net and Places dataset for scene net, as described in Section \ref{sec:oscnn}. In order to make the deep-learned features more discriminative for the task of event recognition, we then fine tune the network parameters on the cultural event recognition dataset.

The network weights are learnt using the mini-batch stochastic gradient descent with momentum (set to 0.9). At each iteration, a mini-batch of $256$ samples is constructed by randomly sampling. During training phase, all the images are resized to $256 \times 256$, and a $224 \times 224$ or $227 \times 227$ sub-image is randomly cropped from the image. They are then manipulated with a random horizontal flipping. The dropout ratio for fully-connected layer is set as $0.5$. To overcome the issues of over-fitting, we set the learning rate of hidden layers as $10^{-2}$ times of final layer. The learning rate is initially set to $10^{-2}$, and decreased according to a fixed schedule: decreasing to $10^{-3}$ after 1.4K iterations, to $10^{-4}$ after 2.8K iterations, and training stopped at 4.2K iterations.

During testing phase, we resort to a multi-view voting method \cite{KrizhevskySH12} to classify each image. Like training procedure, we resize each testing image into $256 \times 256$. For each CNN, we obtain $10$ inputs by cropping and flipping four corners and the center of the image. The score of this CNN for this image can be obtained by averaging the scores across these crops. The scores from multiple object and scene nets are combined using late fusion.

\subsection{Experimental results}
\begin{figure*}[h]
  \includegraphics[width=1\textwidth]{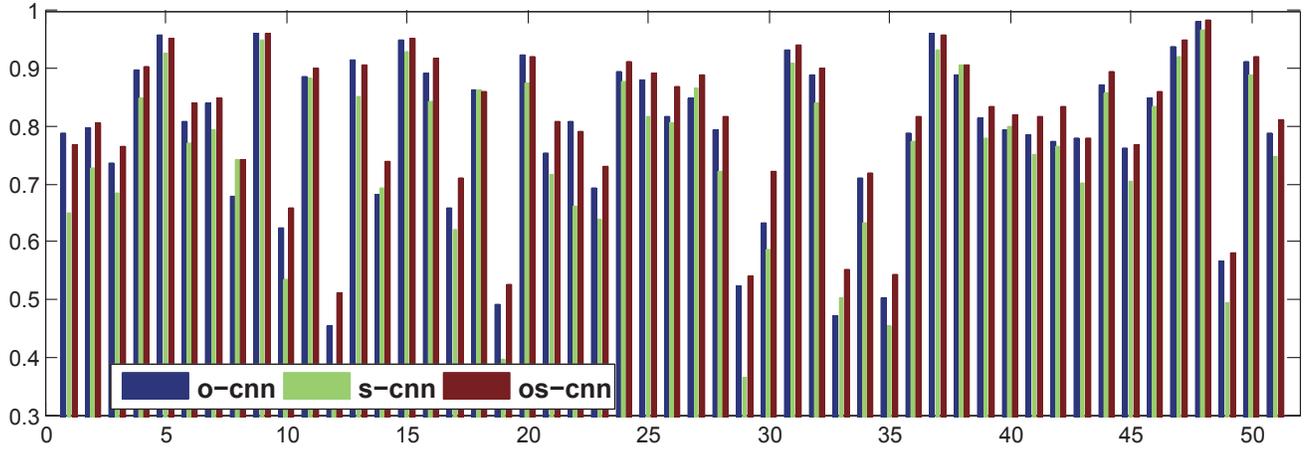}
  \vspace{-0.5cm}
  \caption{Results of object net (o-cnn), scene net (s-cnn), and OS-CNN. We plot the average precision (AP) values for the $50$ classes and the last column indicates the mean AP (mAP) over these classes.}
  \vspace{-0.5cm}
  \label{fig:result1}
\end{figure*}

\textbf{Effectiveness of OS-CNN.} First, we measure the performance of separate object and scene nets. Three scenarios are considered: (i) only using object net, (ii) only using scene net, (ii) using OS-CNN. For each setting, we use the deep network architecture: ClarifaiNet for object net and AlexNet for scene net. The results are shown in Figure \ref{fig:result1}. From these results, object net outperforms scene net for the task of event recognition (mAP $78.8\%$ vs. $74.8\%$). It is also clear that fusion of object and scene nets helps to improve the performance to $81.1\%$. This result indicates there exists complementary property between object and scene nets for event recognition.

\begin{figure*}[h]
  \includegraphics[width=1\textwidth]{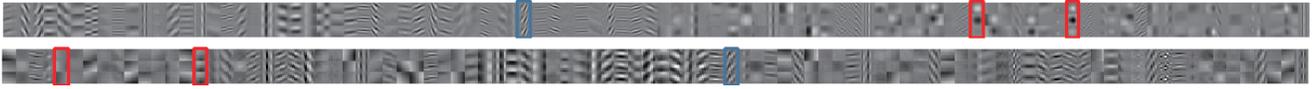}
  \vspace{-0.5cm}
  \caption{The filters learned in first layer of object net and scene net. Blue box indicates similar filters shared by two CNNs, and red boxes denote the filters only learned by a single CNN.}
  \vspace{-0.5cm}
  \label{fig:filter}
\end{figure*}

In order to further investigate this complementarity, we visualize the filters of first convolutional layer of object and scene nets in Figure \ref{fig:filter}. There are 96 filters in the first convolutional layers. We observe that both nets may learn some common filters indicates by the blue box. Meanwhile, some filters indicated by red boxes are only learned by a single net. Therefore, object net and scene net may capture common patterns such edges, but also extract complementary information with different filters.

\textbf{Evaluation of different architectures.} Second, we investigate the performance of CNNs with different architectures and design three settings: (i) CNN with deep architecture (AlexNet or ClarifaiNet), (ii) CNN with very-deep architecture (GoogLeNet), (iii) combination of both deep architecture and very-deep architecture. We conduct this comparative study for both object and scene nets.

\begin{figure*}[h]
  \includegraphics[width=1\textwidth]{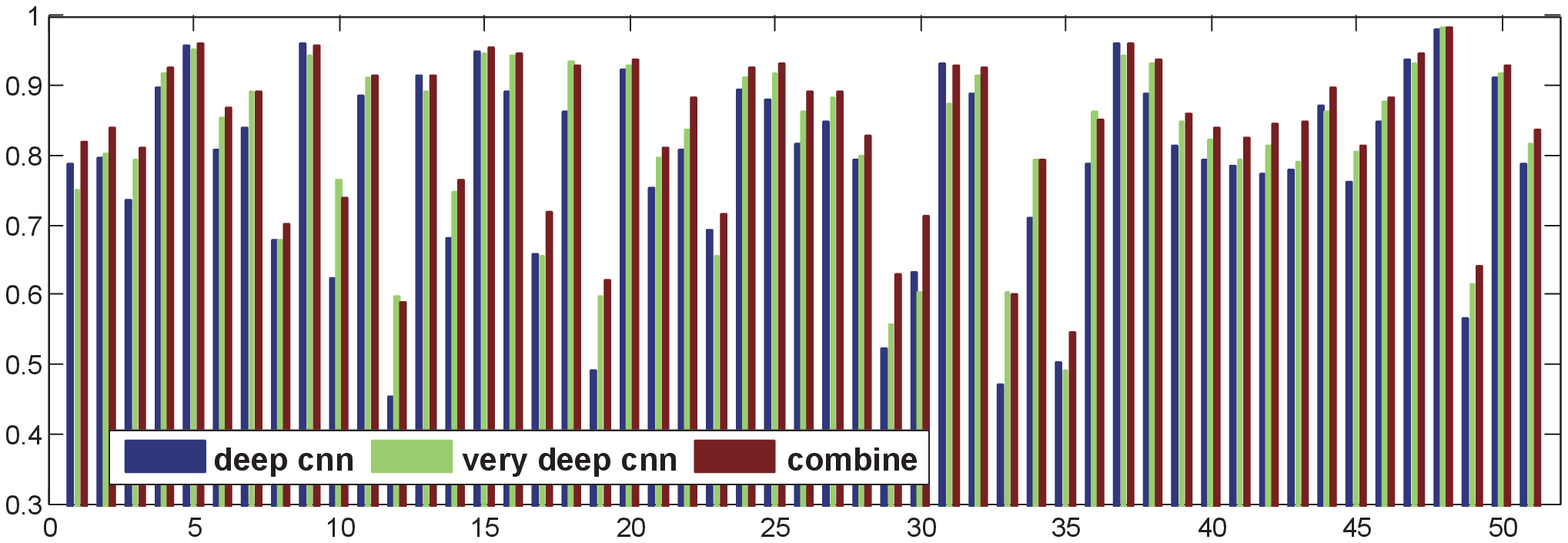}
  \vspace{-0.5cm}
  \caption{Results of \textbf{object net} using different architectures. We plot the average precision (AP) values for the $50$ classes and the last column indicates the mean AP (mAP) over these classes.}
  \vspace{-0.5cm}
  \label{fig:result2}
\end{figure*}

\begin{figure*}[h]
  \includegraphics[width=1\textwidth]{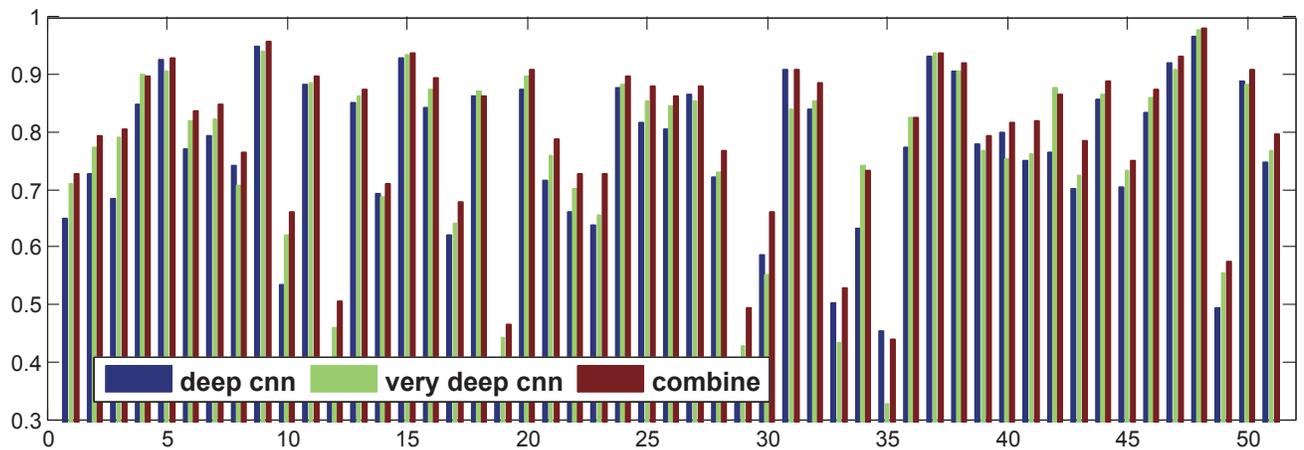}
  \vspace{-0.5cm}
  \caption{Results of \textbf{scene net} using different architectures. We plot the average precision (AP) values for the $50$ classes and the last column indicates the mean AP (mAP) over these classes.}
  \vspace{-0.5cm}
  \label{fig:result3}
\end{figure*}

The results of object net and scene net are shown in Figure \ref{fig:result2} and Figure \ref{fig:result3} respectively. From these results, it is clear that deeper architecture obtain better performance for event recognition, no matter object net or scene net, which agrees with the findings in object recognition. The very-deep architecture outperforms deep architecture by about $3\%$. At the same time, we observe that the fusion of different architectures can help to further boost the recognition performance (about $2\%$ improvement).

\textbf{Challenge approach and results.} Based on the numerical evaluation and analysis above, we conclude that (i) object net is better than scene net, (ii) very-deep architecture outperforms deep architecture, (iii) fusion of multiple CNNs from different visual cues (object and scene) with different architectures (deep and very-deep) contributes to performance improvement. Hence, we introduce another object net with very-deep architecture into our OS-CNN framework. We pre-train a 19-layer VGGNet on the ImageNet dataset and fine tune network weights on the training dataset of cultural event recognition. Totally, our challenge solution is composed of five-stream CNNs pre-trained with different datasets (ImageNet or Places) equipped with different network architectures. The challenge results are shown in Table \ref{tbl:challenge}. We see that our method obtains the best performance and significantly outperforms the second place by nearly $10\%$.

\begin{table}
\centering
\begin{tabular}{|c|c|c|}
\hline
~~~Rank~~~ & ~~~~Team~~~~ & ~~~Score~~~\\
\hline
\hline
1 & \textbf{MMLAB (Ours)} & 85.5\% \\
2 & UPC-STP & 76.7\% \\
3 & MIPAL\_SNU & 73.5\% \\
4 & SBU\_CS & 61.0\% \\
5 & MasterBlaster & 58.2\% \\
6 & Nyx & 31.9\% \\
\hline
\end{tabular}
\vspace{1mm}
\caption{Comparison the performance of our five-stream CNN with that of other team. Our result is significantly better than others.}
\vspace{-5mm}
\label{tbl:challenge}
\end{table}


\section{Conclusions}
\label{sec:conclusion}
This paper has presented an effective method for cultural event recognition from still images. We utilize the deep CNNs for this task and propose a new architecture, called \emph{Object-Scene Convolutional Neural Network} (OS-CNN). This architecture is decomposed into object net and scene net, which extract useful information for event understanding from the perspective of objects and scene content, respectively. Meanwhile, we consider different network structures for OS-CNN and conduct a comparative study of deep CNN and very-deep CNN for event recognition. We show that deeper architecture is also helpful in the task of event recognition from still images, and the combination of different architectures is able to boost performance. In practice, based on our proposed OS-CNN and comparative study, we design a five-stream CNN for the track of cultural event recognition at the ChaLearn LAP challenge 2015. In the future, we may consider jointly optimizing the object and scene nets and incorporating more visual cues for event understanding.

\section*{Acknowledgement}
 This work is supported by a donation of Tesla K40 GPU from NVIDIA corporation. Limin Wang is supported by Hong Kong PhD Fellowship. Yu Qiao is supported by National Natural Science Foundation of China (91320101, 61472410), Shenzhen Basic Research Program (JCYJ20120903092050890, JCYJ20120617114614438, JCYJ20130402113127496), 100 Talents Program of CAS, and Guangdong Innovative Research Team Program (No.201001D0104648280).

{
\bibliographystyle{ieee}
\bibliography{lap}
}

\end{document}